\documentclass[a4paper,fleqn]{cas-dc}

\usepackage[numbers]{natbib}
\usepackage{adjustbox}
\usepackage{graphicx}
\usepackage{subcaption}
\usepackage[utf8]{inputenc}
\usepackage[T1]{fontenc}

\def\tsc#1{\csdef{#1}{\textsc{\lowercase{#1}}\xspace}}
\tsc{WGM}
\tsc{QE}
\tsc{EP}
\tsc{PMS}
\tsc{BEC}
\tsc{DE}
\let\printorcid\relax

\begin{document}
\let\WriteBookmarks\relax
\def\floatpagepagefraction{1}
\def\textpagefraction{.001}

\shorttitle{Improving Precipitation Nowcasting with Multi-Quantile Regression}

\shortauthors{G. van Nieuwkoop, S. Mehrkanoon}

\title [mode = title]{Beyond MSE: Improving Precipitation Nowcasting with Multi-Quantile Regression}                      
% \tnotemark[1,2]

% \tnotetext[1]{This document is the results of the research
   % project funded by the National Science Foundation.}

% \tnotetext[2]{The second title footnote which is a longer text matter
%    to fill through the whole text width and overflow into
%    another line in the footnotes area of the first page.}

\author{Gijs van Nieuwkoop}
\ead{gijsvannieuwkoop@gmail.com}

%\affiliation{Department of Information and Computing Sciences, Utrecht University, Utrecht, The Netherlands}

\address{Department of Information and Computing Sciences, Utrecht University, Utrecht, The Netherlands}

\author{Siamak Mehrkanoon}

\cormark[1]
\ead{s.mehrkanoon@uu.nl} % use Siamak's correct email address here

\cortext[cor1]{Corresponding author}

%%%%%%%  test %%%%%%

%%%%%%%%%%%%%%%%%%%%

\begin{abstract}
Deep-learning precipitation nowcasting models are often optimized using pointwise losses such as mean squared error or mean absolute error, which can lead to overly smooth forecasts and poor representation of heavy rainfall. This study investigates whether the predictive performance of an established deterministic nowcasting architecture can be improved by reformulating training as a multi-quantile regression problem. Using SmaAt-UNet as a core model, we compare MSE, MAE, and multi-quantile pinball-loss training on radar precipitation nowcasting over the Netherlands. The results show that multi-quantile training improves the central deterministic forecast, decreasing test-set MSE by 8.6\% compared to a model trained using MSE, while also producing upper-quantile outputs that are useful for risk-sensitive prediction of heavy precipitation. These findings suggest that quantile regression provides a simple alternative to standard pointwise losses without requiring a new architecture or generative sampling procedure. The implementation of our models and training setup is available on 
\href{https://github.com/gijsvn/Multi-Quantile-Precipitation-Nowcasting}{GitHub}.
\end{abstract}

% \includegraphics{figs/grabs.pdf}

% \begin{highlights}
% \item Research highlights item 1
% \item Research highlights item 2
% \item Research highlights item 3
% \end{highlights}

% Keywords
\begin{keywords}
Precipitation Nowcasting \sep 
Deep Neural Networks \sep
Quantile Regression \sep
SmaAt-UNet
\end{keywords}

\maketitle

\section{Introduction}

Accurate short-term precipitation forecasting is important for weather-sensitive decision making, especially when localized high-intensity rainfall may cause large impacts. In the first minutes to hours ahead, such forecasts are commonly referred to as \emph{nowcasts}, and are typically based on recent radar, satellite or station observations. Because precipitation evolves rapidly and has strong spatial and temporal structure, nowcasting has become a natural application area for deep learning. Most deep-learning nowcasting models formulate the task as spatiotemporal sequence prediction, where past precipitation maps are used to predict one or more future maps \cite{mehrkanoon2019deep, trebing2020wind, wang2017predrnn, fernandez2022deep, stanczyk2021deep}. A broad range of architectures has been proposed, including recurrent convolutional models, encoder--decoder networks, attention-based U-Nets and graph-based models \citep{shi2015convolutional, shi2017deep, trebing2021smaatunet, vatamany2025graph, yang2022aa, kaparakis2023wf, fernandez2021broad, espeholt2022deeplearning, sonderby2020metnet}. However, forecast quality is determined not only by the architecture, but also by the objective used to train it.

Many deterministic nowcasting models are trained with pointwise losses such as mean squared error or mean absolute error. These objectives are simple and stable, but they tend to favor conditional-average predictions when several future precipitation evolutions are plausible. In radar nowcasting, this can lead to spatial smoothing, loss of fine-scale structure and underprediction of rare but meteorologically important heavy rainfall. The problem is amplified by the strong imbalance of precipitation data, where most pixels contain no or weak precipitation and intense rainfall occupies only a small fraction of the domain. Weighted and balanced losses can partly address this by emphasizing heavier precipitation \citep{shi2017deep, ravuri2021skilful}, but improving high-threshold skill often comes at the cost of degraded performance elsewhere.

Probabilistic and generative nowcasting models address uncertainty more directly by producing multiple plausible future precipitation maps. These approaches can generate sharper and more realistic precipitation structures than conventional deterministic models, particularly at longer lead times and for heavier rainfall \citep{ravuri2021skilful, zhang2023nowcastnet, leinonen2023ldcast}. However, they are often more complex to train and sample from, and may be less convenient when a single actionable forecast is required. Quantile regression provides a simpler alternative by estimating selected conditional quantiles of the predictive distribution rather than only the conditional mean \citep{koenker1978regression, koenker2005quantile}. In neural networks, this can be implemented using the asymmetric absolute-error, or pinball loss \cite{wang2022comprehensive}. The median quantile can be used as a central deterministic forecast, while higher quantiles provide upper-tail estimates that are naturally suited to risk-sensitive prediction of heavy precipitation.

In this study, we test whether an established deterministic nowcasting architecture can be improved by changing the training objective rather than the model design. Using \textit{Small Attention U-Net} (SmaAt-UNet) as a core model \cite{trebing2021smaatunet}, we compare conventional MSE and MAE training with multi-quantile pinball-loss training. The model predicts several future precipitation quantiles, with the median output used as the deterministic forecast. Our goal is to isolate the effect of multi-quantile training and assess whether it improves central forecast accuracy while also producing upper-tail outputs for risk-sensitive prediction of heavier precipitation.

\section{Related work}\label{sec:related work}

Deep-learning precipitation nowcasting has been approached through several architectural families. Early recurrent convolutional models such as ConvLSTM and TrajGRU formulated radar nowcasting as spatiotemporal sequence prediction, with TrajGRU introducing learned location-variant connections to better represent precipitation motion \citep{shi2015convolutional, shi2017deep}. Fully convolutional encoder-decoder models later became popular for mapping past radar observations directly to future precipitation maps; RainNet showed that U-Net-like models can outperform optical-flow baselines, while also illustrating the smoothing tendency of deterministic CNN forecasts \citep{ayzel2020rainnet}. Subsequent models improved efficiency or scale, including SmaAt-UNet with attention and depthwise-separable convolutions, and larger neural weather models such as MetNet and MetNet-2 for longer horizons and larger spatial contexts \citep{trebing2021smaatunet, sonderby2020metnet, espeholt2022deeplearning}. While these studies primarily focus on architecture design, the present work focuses on how an established architecture should be optimized.

A second line of work studies training objectives that better account for the imbalance and heavy-tailed nature of precipitation. Standard pixel-wise losses such as MSE and MAE are easy to optimize, but can underemphasize rare high-intensity rainfall and encourage forecasts that perform well on average while missing practically important events. Balanced variants of MSE and MAE address this by assigning larger weights to higher precipitation intensities \citep{shi2017deep}, and related work has explored hybrid weighting, high-intensity-oriented losses and focal losses for difficult future frames \citep{cao2022hybrid, ko2022effective, ma2022focal}. Other objectives incorporate perceptual or structural criteria, such as SSIM-based image-quality losses, to reduce the mismatch between pixel-wise error minimization and visually or meteorologically useful forecasts \citep{tran2019computer}. These studies show that loss functions strongly affect nowcasting skill, but most remain deterministic regression objectives or threshold-weighted variants thereof.

Probabilistic nowcasting instead represents forecast uncertainty directly. Deep generative models, including adversarial and diffusion-based approaches, can produce sharper and more realistic future radar maps than conventional deterministic models \citep{ravuri2021skilful, zhang2023nowcastnet, leinonen2023ldcast}, but they are often more complex to train and may require sampling or post-processing when a single forecast is needed. Quantile regression provides a lighter-weight alternative, allowing models to estimate selected conditional quantiles rather than only a conditional mean. Quantile regression is well established in statistics and has been used for neural-network-based probabilistic forecasting and weather post-processing \citep{koenker1978regression, taylor2000quantile, rasp2018neural}. Direct applications to precipitation nowcasting remain comparatively limited, although recent work has begun to use quantile-regression neural networks and pinball losses for probabilistic prediction and uncertainty estimation \citep{schaumann2024generating, baron20253d}. In contrast, the present study evaluates multi-quantile training as a practical extension of deterministic nowcasting. The contributions are twofold: (i) We show how multi-quantile training can be used as an auxiliary optimization strategy, where non-central quantiles provide additional learning signals that may improve generalization beyond a single deterministic target. (ii) We show how higher-quantile outputs obtained from the same training procedure can serve as risk-sensitive predictors for rare heavy and extreme precipitation events.

\begin{figure*}
	\centering
    \adjustbox{center}{\includegraphics[width=\textwidth]{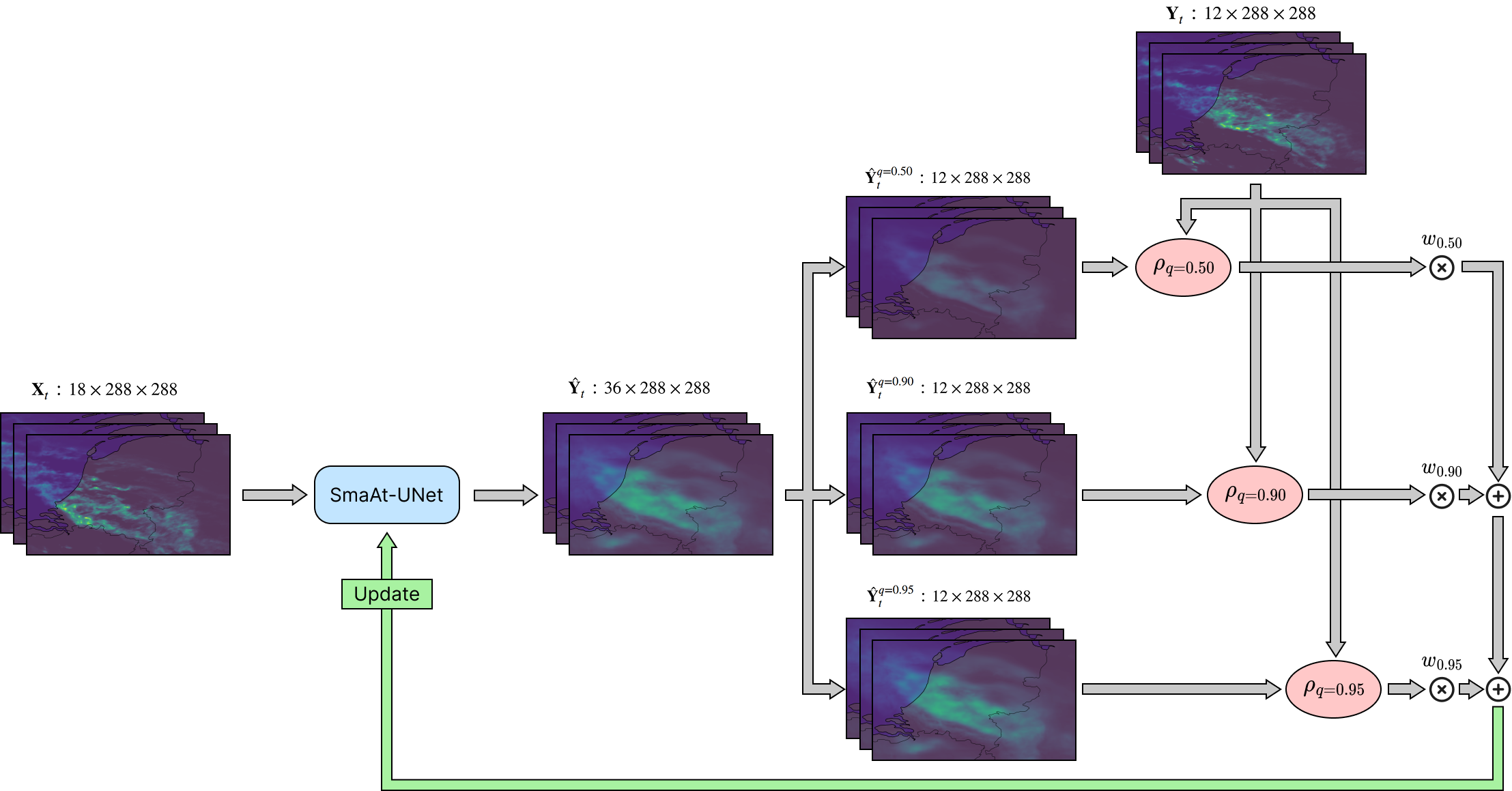}}
	\caption{Schematic overview of the multi-quantile training setup, with pinball losses $\rho_q$ and corresponding weights $w_q$ for quantiles $q \in Q$.}
	\label{fig:training_scheme}
\end{figure*}

\section{Methods}\label{sec:methods}

\subsection{Prediction task and data representation}

We formulate precipitation nowcasting as sequence-to-sequence prediction. Let $\mathbf{X}_t = (x_{t-m+1}, \ldots, x_t)$ denote $m$ past radar precipitation maps and let $\mathbf{Y}_t = (y_{t+1}, \ldots, y_{t+L})$ denote the corresponding sequence of $L$ future maps. A model with parameters $\theta$ learns a mapping $f_\theta: \mathbf{X}_t \mapsto \hat{\mathbf{Y}}_t$. In the deterministic setting, $\hat{\mathbf{Y}}_t \in \mathbb{R}^{L \times H \times W}$, with one predicted map per lead time. In the quantile-regression setting, the output is expanded to $\hat{\mathbf{Y}}_t \in \mathbb{R}^{L \times |\mathcal{Q}| \times H \times W}$, with one prediction for each lead time and quantile $q \in \mathcal{Q}$. All model variants are evaluated against the same observed future sequence; for the quantile model, the $q=0.5$ output is used as the central deterministic forecast.

\subsection{Base architecture}\label{sec:architecture}

All experiments use SmaAt-UNet \cite{trebing2021smaatunet} as the core nowcasting architecture. SmaAt-UNet is a lightweight U-Net-based encoder-decoder model that combines attention mechanisms with depthwise-separable convolutions. To isolate the effect of the training objective, the same backbone is used across all experiments; only the final output dimensionality is changed. The MSE- and MAE-trained models output \(L\) channels, corresponding to one precipitation map per lead time, whereas the quantile-regression model outputs \(L|\mathcal{Q}|\) channels, corresponding to one map for each lead-time-quantile pair, as illustrated in Figure~\ref{fig:training_scheme}.

\subsection{Training Objectives}

\subsubsection{Multi-quantile pinball loss}

For the quantile-regression models, the network predicts multiple conditional quantiles of the future precipitation distribution. Instead of producing a single precipitation estimate for each forecast lead time and grid cell, the model outputs one prediction $\hat{y}^{(q)}$ for each quantile $q \in \mathcal{Q}, 0 < q < 1$, as illustrated in Figure~\ref{fig:training_scheme}. A predicted quantile $\hat{y}^{(q)}$ estimates the value below which the future precipitation intensity is expected to fall with probability $q$, conditional on the input sequence. Thus, the $q=0.50$ output represents the conditional median, while higher quantiles represent upper-tail precipitation estimates.

Throughout this work, we use $\mathcal{Q}=\{0.5,0.9,0.95\}$. The median output is used as the central deterministic forecast when comparing with the MSE- and MAE-trained baselines, while the $q=0.90$ and $q=0.95$ outputs provide increasingly conservative upper-tail estimates for risk-sensitive prediction of heavy and extreme precipitation. This combines a central estimate with two upper quantiles suited to the zero-inflated and heavy-tailed nature of precipitation. The quantile set was chosen empirically, though further optimization of these levels may yield stronger variants.

Quantile regression was optimized using the pinball loss. For a target precipitation value $y$, predicted quantile $\hat{y}^{(q)}$, and error $e = y - \hat{y}^{(q)}$, the pinball loss is defined as follows:
\begin{equation}
\begin{split}
    \rho_q(y, \hat{y}^{(q)})
    &= \max(qe, (q - 1)e) \\
    &=
    \begin{cases}
        q |e|, & e \geq 0, \\
        (1-q)|e|, & e < 0.
    \end{cases}
\end{split}
\label{eq:pinball_loss}
\end{equation}

\begin{figure}
	\centering
    \adjustbox{center}{\includegraphics[width=0.5\textwidth]{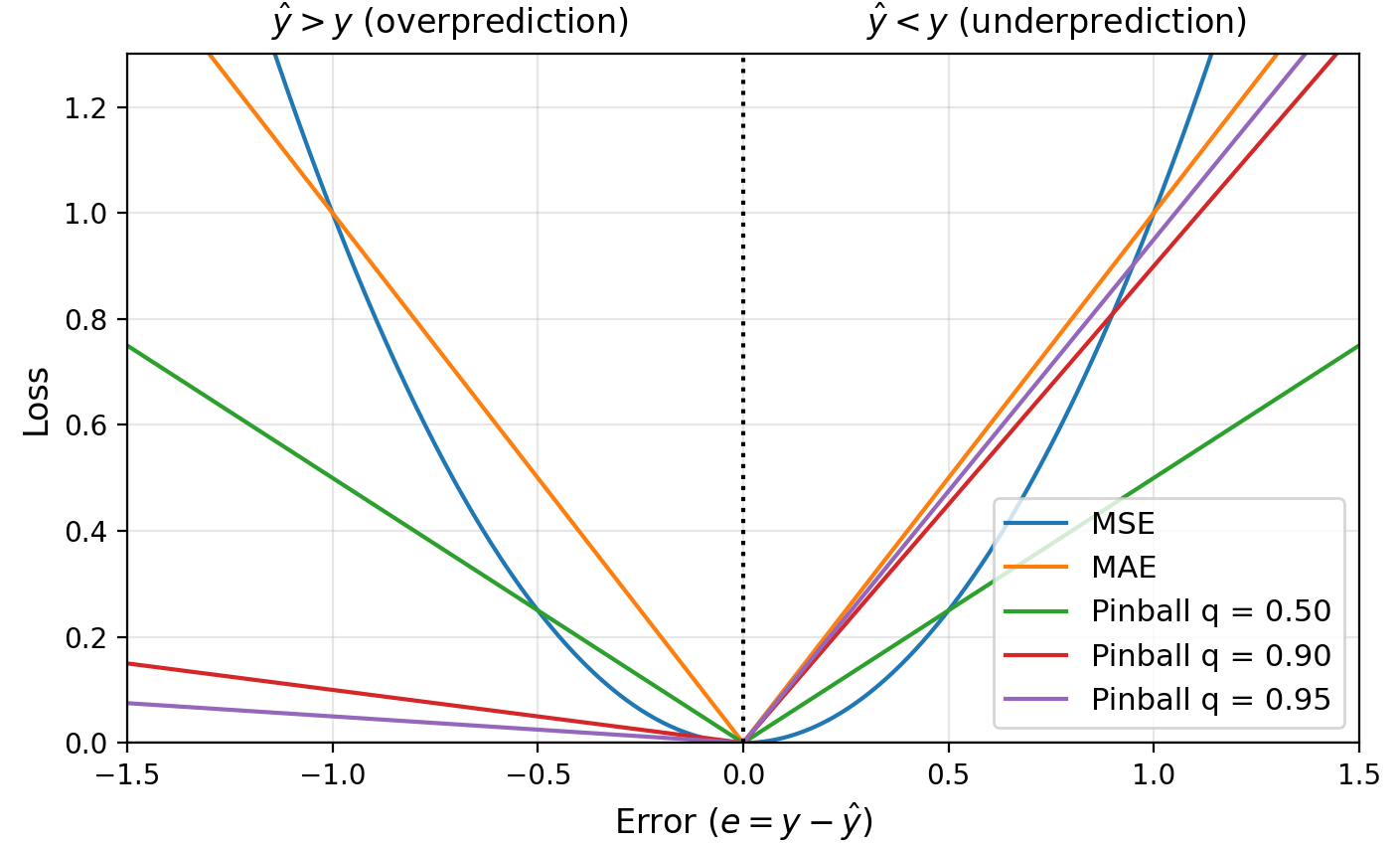}}
	\caption{MSE, MAE, and pinball losses, visualized as functions of the error $e=y - \hat{y}$.}
	\label{fig:loss_visualization}
\end{figure}

This loss is asymmetric unless $q=0.50$, as illustrated in Figure~\ref{fig:loss_visualization}. For the median quantile, the pinball loss is proportional to MAE and has the same optimum, namely the conditional median. For higher quantiles, underprediction is penalized more strongly than overprediction, causing the learned prediction to shift upward relative to the central estimate and represent upper-tail precipitation behavior.

Only one future precipitation sequence is observed for each input sequence, so the model does not receive separate target maps for different quantiles. Instead, each quantile output is compared with the same observed target, but through a different asymmetric loss. Across many training examples, this encourages the $q=0.50$ output to approach the conditional median, while the $q=0.90$ and $q=0.95$ outputs are pushed toward values that are exceeded less often.

The total multi-quantile loss applies the pinball loss separately to each predicted quantile and combines the resulting losses using quantile-specific weights. Let $w_q$ denote the weight assigned to quantile $q$. For a batch of predictions, the implemented objective is as follows:
\begin{equation}
\mathcal{L}_{\mathrm{Q}}
=
\frac{1}{B}
\sum_{b=1}^{B}
\sum_{q \in \mathcal{Q}}
w_q
\sum_{\ell=1}^{L}
\sum_{i=1}^{H}
\sum_{j=1}^{W}
\rho_q
\left(
y_{b,\ell,i,j},
\hat{y}^{(q)}_{b,\ell,i,j}
\right),
\label{eq:multi_quantile_loss}
\end{equation}
where $B$ is the batch size, $L$ is the number of forecast lead times, and $H \times W$ is the spatial grid size. Figure~\ref{fig:training_scheme} illustrates this procedure: each quantile-specific pinball loss is weighted by $w_q$, and the weighted losses are summed into a single training objective.

\begin{figure}
	\centering
    \adjustbox{center}{\includegraphics[width=\linewidth]{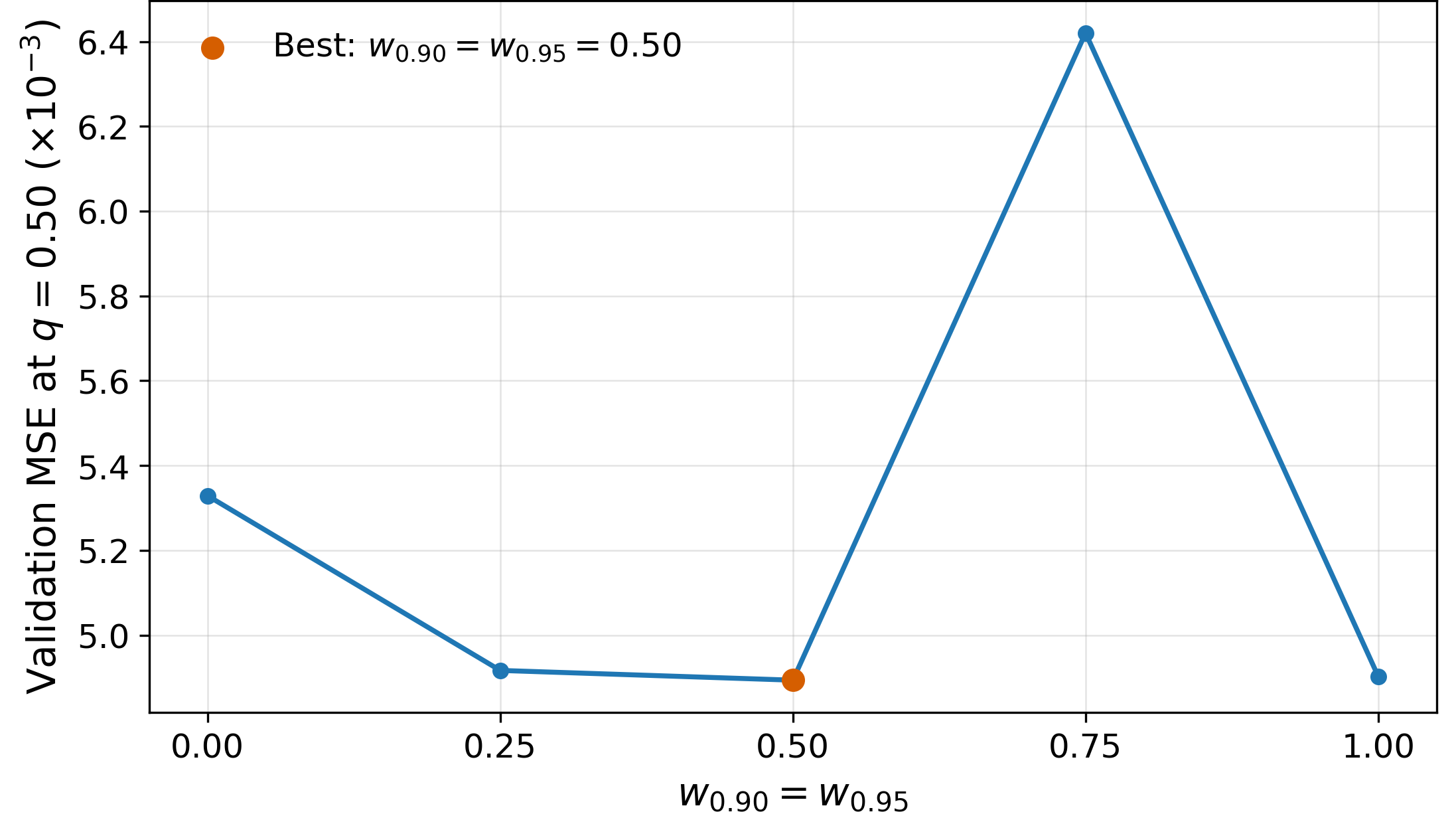}}
	\caption{One-dimensional grid search over the shared upper-quantile loss weight \(w_{0.90}=w_{0.95}\), with \(w_{0.50}\) fixed at 1.0.}
	\label{fig:quantile_weight_gs}
\end{figure}

In experiments, the quantile-loss weights were set to \(w_{0.50}=1.0\), \(w_{0.90}=0.5\), and \(w_{0.95}=0.5\). These values were selected using a one-dimensional grid search in which \(w_{0.50}\) was fixed at 1.0 and the upper-quantile weights were constrained to be equal, i.e. \(w_{0.90}=w_{0.95}\). The validation MSE of the median output was used as the selection criterion. The full grid-search results are shown in Figure~\ref{fig:quantile_weight_gs}. The selected configuration retains the largest weight on the central estimate while still incorporating training feedback from the higher quantiles.

Since all quantile outputs share the same SmaAt-UNet backbone, the upper-quantile losses not only train separate output maps but also influence the shared representation used by the median forecast. Multi-quantile training can therefore be interpreted as a form of auxiliary training, where the model is encouraged to learn features that support both central precipitation prediction and risk-sensitive upper-tail estimation.

\subsubsection{Other objectives}

% For comparison with the multi-quantile regression objective, models were also trained using mean squared error (MSE) and mean absolute error (MAE). These objectives produce a single predicted precipitation map for each forecast lead time, rather than multiple quantile estimates. Unlike the upper-quantile pinball losses, MSE and MAE are symmetric with respect to the sign of the error: an overprediction and an underprediction of the same magnitude receive the same penalty, as shown in Figure~\ref{fig:loss_visualization}.

% For a target precipitation value $y$ and model prediction $\hat{y}$, the mean squared error is defined as:

% \begin{equation}
% \mathcal{L}_{\mathrm{MSE}} =
% \frac{1}{B}
% \sum_{b=1}^{B}
% \sum_{\ell=1}^{L}
% \sum_{i=1}^{H}
% \sum_{j=1}^{W}
% (y_{b,\ell,i,j} - \hat{y}_{b,\ell,i,j})^2.
% \label{eq:mse_loss}
% \end{equation}
% The mean absolute error is defined as
% \begin{equation}
% \mathcal{L}_{\mathrm{MAE}} =
% \frac{1}{B}
% \sum_{b=1}^{B}
% \sum_{\ell=1}^{L}
% \sum_{i=1}^{H}
% \sum_{j=1}^{W}
% |y_{b,\ell,i,j} - \hat{y}_{b,\ell,i,j}|.
% \label{eq:mae_loss}
% \end{equation}

% MSE and MAE were included because they are widely used pointwise regression losses for deterministic precipitation nowcasting. MSE penalizes larger errors more strongly because the error term is squared, while MAE applies a linear penalty to all absolute errors. Comparing these objectives with the multi-quantile pinball loss makes it possible to evaluate the effect of the training objective while keeping the model architecture and prediction task fixed. 

For comparison, we also trained deterministic SmaAt-UNet models using mean squared error (MSE) and mean absolute error (MAE), both widely used pointwise regression losses, which therefore serve as natural deterministic benchmarks. These models output a single precipitation map per forecast lead time, rather than multiple quantile estimates. For target precipitation values $y$ and predictions $\hat{y}$, the losses are defined as follows:

\begin{equation}
\mathcal{L}_{\mathrm{MSE}}=
\frac{1}{B}
\sum_{b=1}^{B}
\sum_{\ell=1}^{L}
\sum_{i=1}^{H}
\sum_{j=1}^{W}
(y_{b,\ell,i,j} - \hat{y}_{b,\ell,i,j})^2, \\
\label{eq:mse_loss}
\end{equation}
\begin{equation}
\mathcal{L}_{\mathrm{MAE}}=
\frac{1}{B}
\sum_{b=1}^{B}
\sum_{\ell=1}^{L}
\sum_{i=1}^{H}
\sum_{j=1}^{W}
|y_{b,\ell,i,j} - \hat{y}_{b,\ell,i,j}|.
\label{eq:mae_loss}
\end{equation}

Unlike upper-quantile pinball losses, MSE and MAE are symmetric with respect to the sign of the error, as shown in Figure~\ref{fig:loss_visualization}. MSE penalizes large errors quadratically, whereas MAE applies a linear penalty to absolute errors. Comparing these benchmarks with the multi-quantile pinball loss allows us to isolate the effect of the training objective while keeping the architecture and prediction task fixed.

\subsection{Evaluation metrics}

Model performance was evaluated using regression metrics and threshold-based event metrics. Regression performance is reported using test-set mean squared error (MSE) and mean absolute error (MAE), following the pointwise definitions in Equations \ref{eq:mse_loss} and \ref{eq:mae_loss}; for evaluation, however, the metrics were averaged over all grid cells and lead times, rather than batch size.

For event-based evaluation, predicted and observed precipitation maps were converted to binary maps using precipitation-rate thresholds. The resulting true positives (TP), true negatives (TN), false positives (FP) and false negatives (FN) were used to compute the critical success index (CSI), probability of detection (POD), false alarm rate (FAR) and Matthews correlation coefficient (MCC):

\begin{align}
\text{CSI} &= \frac{\mathrm{TP}}{\mathrm{TP}+\mathrm{FP}+\mathrm{FN}}\label{eq:csi} \\[4pt]
\text{POD} &= \frac{\mathrm{TP}}{\mathrm{TP}+\mathrm{FN}}\label{eq:pod} \\[4pt]
\text{FAR} &= \frac{\mathrm{FP}}{\mathrm{TP}+\mathrm{FP}}\label{eq:far} \\[4pt]
\text{MCC} &= \frac{\mathrm{TP}\,\mathrm{TN}-\mathrm{FP}\,\mathrm{FN}}{\sqrt{(\mathrm{TP}+\mathrm{FP})(\mathrm{TP}+\mathrm{FN})(\mathrm{TN}+\mathrm{FP})(\mathrm{TN}+\mathrm{FN})}}\label{eq:mcc}.
\end{align}

The critical success index is used as the primary event-based metric, as is common in precipitation nowcasting, because it measures rainfall-event overlap while ignoring the many true negatives in imbalanced precipitation maps. The remaining metrics provide complementary information: probability of detection measures sensitivity to observed events, false alarm rate quantifies overprediction, and the Matthews correlation coefficient summarizes all four confusion-matrix entries, remaining informative under strong class imbalance. Together, these metrics give a compact view of different aspects of nowcasting performance, while evaluating them at increasing thresholds assesses performance on increasingly rare and heavy precipitation events.

\section{Experiments}\label{sec:experiments}

To isolate the effect of the optimization objective, all models were trained and evaluated under the same experimental setup, using the same dataset, prediction task, preprocessing, training framework and SmaAt-UNet backbone, with only the loss function and output dimensionality varied.

\subsection{Dataset and preprocessing}

We use the same KNMI radar precipitation dataset as the original SmaAt-UNet study \cite{trebing2021smaatunet}. The dataset contains approximately 420,000 radar precipitation maps over the Netherlands from 2016-2019, recorded at 5-minute intervals on a 288×288 grid with pixel sizes of approximately 1 km$^2$. Pixel values represent 5-minute precipitation accumulations in millimetres; for threshold-based evaluation, these values are converted to precipitation rates in mm/h by multiplying by 12. Given 18 radar maps from the previous 90 minutes, the model predicts the next 12 maps, corresponding to lead times from 5 to 60 minutes at 5-minute resolution. The dataset was split chronologically, using 2016-2018 for training and 2019 for testing, with 10\% of the training samples selected for validation using seasonally stratified sampling. Pixel values were normalized by the maximum value observed in the training set, and the same normalization factor was applied to the validation and test sets. To reduce the dominance of low-rainfall cases, only samples whose final target frame contained non-zero precipitation in at least 50\% of pixels were retained. This yielded 5,734 training samples and 1,557 test samples.

\subsection{Training setup}

% All models were optimized using the Adam optimizer with an initial learning rate of $0.001$ and a batch size of 16. A learning-rate scheduler was used to reduce the learning rate by a factor of $0.1$ whenever the validation loss did not improve for 4 consecutive epochs. Training was terminated when the validation loss did not improve for 15 consecutive epochs. For each model configuration, five independent training runs were performed. This was done to reduce the influence of stochastic effects from random weight initialization and mini-batch ordering. After training, the model checkpoint with the lowest validation loss across the five runs was selected for evaluation on the test set. All training runs were performed on a single NVIDIA H100 GPU with 94 GB of VRAM.

All models were optimized with an Adam optimizer using an initial learning rate of $10^{-3}$ and a batch size of 16. The learning rate was reduced by a factor of $0.1$ after 4 epochs without improvements to validation loss, and training was terminated after 15 such epochs. To reduce sensitivity to random initialization and mini-batch ordering, each configuration was trained five times, after which the checkpoint with the lowest validation loss across runs was evaluated on the test set. All runs were performed on a single NVIDIA H100 GPU with 94 GB of VRAM.

\begin{table}
\centering
\footnotesize
\caption{Test-set regression performance averaged over forecast lead time. Quantile-loss results use the median prediction ($q=0.50$). Bold and underlined values denote the best and second-best scores, respectively.}
\begin{tabular}{l c c}
\toprule
\textbf{Training Loss} & \textbf{Test Set MSE} & \textbf{Test Set MAE}     \\
\midrule
MSE                    & \underline{0.0151}    & 0.0424             \\
MAE                    & 0.0161                & \underline{0.0348} \\
Quantile               & \textbf{0.0138}       & \textbf{0.0345}    \\
\bottomrule
\end{tabular}
\label{tab:reg_result_table}
\end{table}

\section{Results and discussion}\label{sec:results}

\begin{figure*}
	\centering
    \adjustbox{center}{\includegraphics[width=1\textwidth]{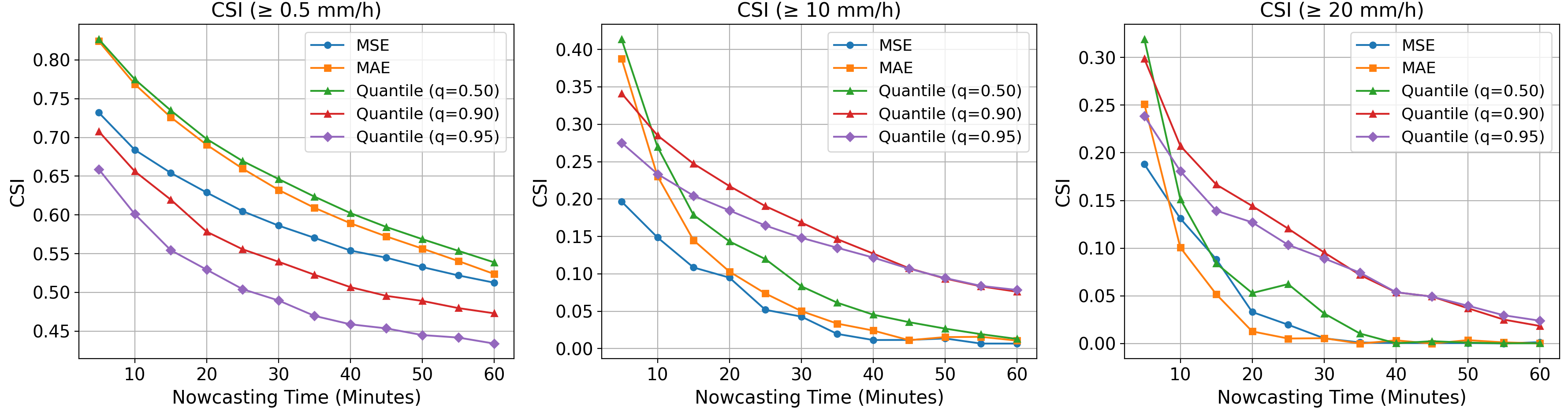}}
	\caption{CSI as a function of forecast lead time for precipitation thresholds of $0.5$, $10$ and $20$ mm/h.}
	\label{fig:csi_per_forecasting_time}
\end{figure*}

\begin{table}[!t]
\centering
\footnotesize
\setlength{\tabcolsep}{4pt}
\renewcommand{\arraystretch}{1}
\caption{Classification performance averaged over forecast lead time for different precipitation thresholds. Arrows indicate whether higher or lower values are better; bold and underlined values denote the best and second-best scores for each threshold.}
\begin{tabular}{l l c c c c}
\toprule
\textbf{Threshold} & 
\textbf{Loss}         & \textbf{CSI ↑}    & \textbf{POD ↑}    & \textbf{FAR ↓}    & \textbf{MCC ↑}    \\
\midrule
\multirow{5}{*}{$\geq 0.5$ mm/h}
& MSE                 & 0.594             & 0.837             & 0.330             & 0.605             \\
& MAE                 & \underline{0.641} & 0.779             & \underline{0.223} & \underline{0.670} \\
& Quantile$_{q=0.50}$ & \textbf{0.652}    & 0.795             & \textbf{0.223}    & \textbf{0.681}    \\
& Quantile$_{q=0.90}$ & 0.552             & \underline{0.970} & 0.439             & 0.563             \\
& Quantile$_{q=0.95}$ & 0.503             & \textbf{0.985}    & 0.493             & 0.498             \\
\midrule
\multirow{5}{*}{$\geq 10$ mm/h}
& MSE                 & 0.059             & 0.073             & 0.756             & 0.126             \\
& MAE                 & 0.092             & 0.111             & \underline{0.683} & 0.175             \\
& Quantile$_{q=0.50}$ & 0.117             & 0.144             & \textbf{0.646}    & 0.216             \\
& Quantile$_{q=0.90}$ & \textbf{0.174}    & \underline{0.456} & 0.788             & \textbf{0.306}    \\
& Quantile$_{q=0.95}$ & \underline{0.153} & \textbf{0.563}    & 0.831             & \underline{0.304} \\
\midrule
\multirow{5}{*}{$\geq 20$ mm/h}
& MSE                 & 0.039             & 0.053             & 0.867             & 0.078             \\
& MAE                 & 0.036             & 0.044             & 0.865             & 0.073             \\
& Quantile$_{q=0.50}$ & 0.060             & 0.076             & \textbf{0.797}    & 0.115             \\
& Quantile$_{q=0.90}$ & \textbf{0.107}    & \underline{0.282} & \underline{0.859} & \underline{0.197} \\
& Quantile$_{q=0.95}$ & \underline{0.096} & \textbf{0.376}    & 0.890             & \textbf{0.202}    \\
\bottomrule
\end{tabular}
\label{tab:clas_result_table}
\end{table}

% Table~\ref{tab:reg_result_table} reports the regression performance of the SmaAt-UNet models trained with MSE, MAE, and multi-quantile pinball loss. The quantile-regression model achieves the lowest test-set MSE and MAE, with values of 0.0138 and 0.0345, respectively. Relative to the MSE-trained baseline, this corresponds to a reduction of approximately 8.6\% in MSE and 18.6\% in MAE. Compared with the MAE-trained model, the quantile model reduces MSE by 14.3\% and obtains a slightly lower MAE, with a 0.8\% reduction. This is an important result because the median output of the quantile model improves on both deterministic baselines, even though the baselines were directly optimized for the metrics on which they are evaluated.

Table~\ref{tab:reg_result_table} shows that the multi-quantile model achieves the best regression performance, with a test-set MSE of $0.0138$ and MAE of $0.0345$. Relative to the MSE-trained baseline, this corresponds to reductions of $8.6\%$ in MSE and $18.6\%$ in MAE; relative to the MAE-trained baseline, MSE and MAE are reduced by $14.3\%$ and $0.8\%$. Thus, the median output of the quantile model improves baselines even though these were trained directly on the pointwise losses used for evaluation.

% A plausible explanation is that multi-quantile training provides a richer learning signal than single-output deterministic regression. The median forecast is trained alongside upper quantiles, which may encourage the shared backbone to learn representations that better capture the full range of possible precipitation outcomes. In this sense, the upper quantile objectives may act as auxiliary training signals that improve the central forecast. Importantly, these results indicate that reformulating the objective as multi-quantile regression improves the deterministic forecast produced by the same base architecture.

A likely explanation is that the upper-quantile objectives provide auxiliary training signals for the shared SmaAt-UNet backbone. By training the median forecast jointly with upper-tail estimates, the model may learn representations that are more informative than those obtained from a single deterministic target, improving the central nowcast as well as the non-central quantile outputs.

% The threshold-based metrics in Table~\ref{tab:clas_result_table} further show that this improvement is not limited to regression error. At the lowest threshold of 0.5 mm/h, the median output of the quantile model achieves the best CSI, FAR, and MCC values, indicating better general precipitation-event prediction than both deterministic baselines. The higher quantile outputs achieve much higher POD values at this threshold, but this comes with substantially increased FAR. They are therefore not well suited as general-purpose low-threshold precipitation forecasts. Instead, their behaviour is better understood as a deliberate shift toward higher sensitivity.

The event-based results in Table~\ref{tab:clas_result_table} show that these gains are not limited to regression error. At the lowest threshold of $0.5$ mm/h, the quantile median obtains the best CSI, FAR and MCC, indicating stronger general rainfall-event prediction than both deterministic baselines. The $q=0.90$ and $q=0.95$ outputs achieve much higher POD at this threshold, but at the cost of substantially more false alarms, making them less suitable as general-purpose low-threshold forecasts.

% At the higher precipitation thresholds, the value of the quantile formulation becomes even clearer. For events exceeding 10 and 20 mm/h, the median output of the quantile model outperforms the MSE and MAE baselines across all reported event-based metrics. This is one of the strongest results of the study: the same central forecast improves on the baselines in continuous regression performance, threshold-based detection of general precipitation and detection of rare events of heavy and extreme precipitation. This suggests that multi-quantile training does not merely produce useful upper-tail outputs, but also improves the general predictive quality of the central nowcast.

% The upper quantile forecasts provide a second, complementary benefit. At the 10 and 20 mm/h thresholds, the q = 0.90 and q = 0.95 outputs substantially increase POD relative to the central forecast, showing that they detect a much larger fraction of heavy-precipitation events. Although the higher quantiles also increase FAR, the balanced CSI and MCC metrics improve as well, indicating that the additional detections outweigh the extra false alarms under these event-based evaluation criteria. The upper quantiles therefore provide more than a sensitivity-adjusted version of the central forecast: for heavy-precipitation thresholds, they represent a more effective operating point for risk-sensitive prediction. This makes them especially relevant for applications where missed heavy-rainfall events are costly, such as flood warning, traffic management, and other operational high-impact weather settings.

At the higher thresholds, the advantage of quantile training becomes even clearer. For events exceeding $10$ and $20$ mm/h, the quantile median outperforms both deterministic baselines across all reported event-based metrics, while the upper quantiles substantially increase POD and also improve CSI and MCC despite higher FAR. This indicates that the upper quantiles are not merely more sensitive forecasts, but better operating points for risk-sensitive heavy-precipitation prediction, where missed events may be more costly than additional false alarms.

\begin{figure*}
	\centering
    \adjustbox{center}{\includegraphics[width=0.95\textwidth]{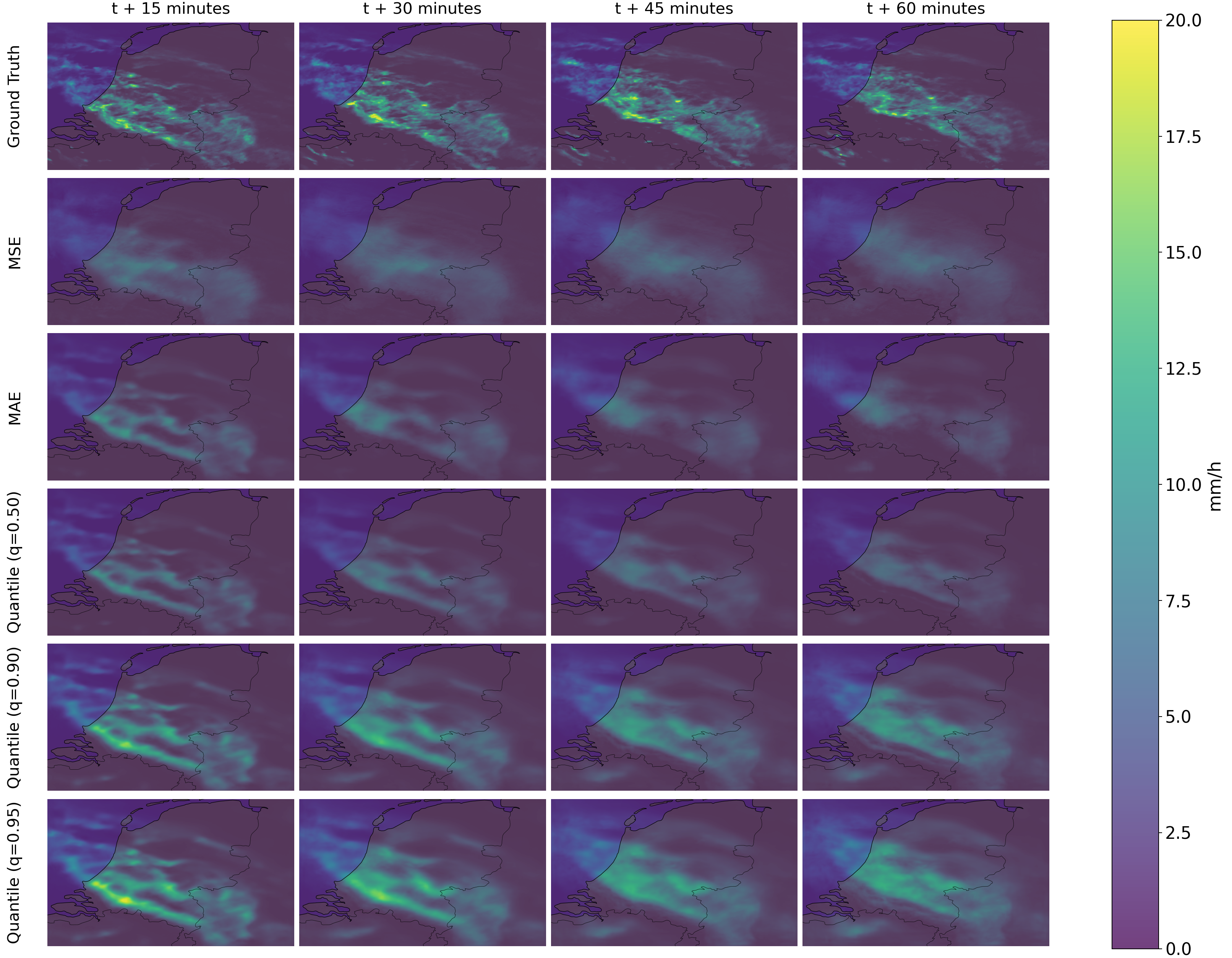}}
	\caption{Qualitative comparison of ground-truth precipitation and model predictions at selected forecast lead times.}
	\label{fig:prediction_visualization}
\end{figure*}

% Figure~\ref{fig:csi_per_forecasting_time} shows how these differences evolve with forecast lead time. Across all models and thresholds, CSI decreases as lead time increases, reflecting the growing uncertainty of precipitation evolution farther into the future. The figure also shows that the advantage of the upper quantile outputs at high precipitation thresholds is most visible at longer lead times. At shorter lead times, the deterministic and median forecasts remain competitive, but as the prediction horizon increases, the higher quantiles retain more skill for detecting heavy rainfall. This supports the interpretation that the upper quantiles are especially useful when uncertainty is larger and high-intensity precipitation becomes harder to localize precisely.

Figure~\ref{fig:csi_per_forecasting_time} shows that CSI decreases with lead time for all models and thresholds, reflecting increasing forecast uncertainty. The advantage of the upper quantiles is clearest at the higher precipitation thresholds and longer lead times, supporting their use when high-intensity precipitation is harder to localize and detection-oriented performance is especially important.
% A case study of predictions for a specific test sample in Figure~\ref{fig:prediction_visualization} helps explain the observed differences in performance metrics. The MSE-trained model produces relatively smooth and diffuse precipitation maps, consistent with the known tendency of squared-error losses to favor conditional-average predictions. The MAE-trained model and the median output of the quantile model preserve more localized precipitation structure. The q = 0.90 and q = 0.95 outputs predict broader and more intense precipitation regions, which explains their increased performance for more intense rain events. The visual results therefore reinforce the quantitative findings: the choice of loss function changes not only scalar performance metrics, but also the spatial character of the generated nowcasts.
The qualitative example in Figure~\ref{fig:prediction_visualization} is consistent with these results. The MSE-trained model produces smoother and more diffuse fields, whereas the MAE-trained model and the quantile median preserve more localized structure. The $q=0.90$ and $q=0.95$ outputs produce broader and more intense precipitation regions, explaining their increased detection of heavy-rainfall events and their higher false-alarm rates.

% Taken together, these results support the central hypothesis of this study. Multi-quantile regression improves the central deterministic forecast while also producing upper-tail forecasts that can be used for risk-sensitive heavy-precipitation prediction. This positions quantile regression as a practical middle ground between standard deterministic losses and more complex probabilistic or generative nowcasting methods. Unlike MSE or MAE training, it provides information about the upper tail of the predictive distribution. Unlike generative approaches, it does so without requiring a new architecture, sampling procedure, or an adversarial- or diffusion-based training framework. The results therefore suggest that loss-function design remains an important and relatively lightweight route for improving deep-learning precipitation nowcasting.

The obtained results show two effects of multi-quantile training: it improves the central deterministic forecast, and it provides upper-tail outputs that are useful for risk-sensitive prediction of heavy precipitation. These gains are obtained by changing only the training objective and output dimensionality, leaving the underlying nowcasting architecture unchanged.

\section{Conclusions}\label{sec:conclusion}

% This study investigated whether precipitation nowcasting performance can be improved by changing the training objective rather than the model architecture. Using SmaAt-UNet as a core model, we compared conventional MSE and MAE training with multi-quantile pinball-loss training, where the median output was used as the central deterministic forecast. The results show that multi-quantile training improves the central forecast across both continuous regression metrics and threshold-based event metrics. In particular, the median output of the quantile-trained model outperformed the MSE- and MAE-trained baselines, indicating that the auxiliary task of predicting upper quantiles can provide a useful training signal for the shared nowcasting model. The higher-quantile outputs further provide risk-sensitive predictors for heavy precipitation. While these forecasts increase false alarms, they substantially improve detection-oriented metrics at high precipitation thresholds, making them useful in applications where missed heavy-rainfall events are especially costly. Overall, these findings suggest that multi-quantile regression is a simple and effective alternative to standard pointwise losses for deterministic precipitation nowcasting. It improves central forecast quality while also providing upper-tail information, without requiring a new architecture, sampling procedure, or fully generative forecasting framework.

This study investigated whether precipitation nowcasting can be improved by changing the training objective rather than the model architecture. Using SmaAt-UNet as core model, we compared MSE and MAE training with multi-quantile pinball-loss training, using the median quantile as the central deterministic forecast. The quantile-trained model improved this central forecast across both regression and event-based metrics, suggesting that upper-quantile prediction provides a useful auxiliary training signal for the shared nowcasting model. The higher quantiles further served as risk-sensitive predictors for heavy precipitation, substantially improving detection-oriented metrics at high precipitation thresholds. Overall, multi-quantile regression provides an alternative to standard pointwise losses, improving deterministic forecast quality while adding upper-tail information without requiring a new architecture, sampling procedure, or fully generative framework.

\bibliographystyle{model1-num-names}
% \bibliographystyle{cas-model2-names}

\bibliography{cas-refs}


\end{document}